\documentclass[a4paper, 10pt, conference]{ieeeconf}      % Use this line for a4 paper
\usepackage[letterpaper, portrait, margin=0.75in]{geometry}
\IEEEoverridecommandlockouts
\usepackage{cite}
\usepackage{amsmath,amssymb,amsfonts}
\usepackage{algorithmic}
\usepackage{graphicx}
\usepackage{physics}
\usepackage[caption=false]{subfig}
\usepackage{textcomp}
\usepackage{xcolor}

\def\BibTeX{{\rm B\kern-.05em{\sc i\kern-.025em b}\kern-.08em
    T\kern-.1667em\lower.7ex\hbox{E}\kern-.125emX}}

\begin{document}

\title{\LARGE \bf SharpSLAM: 3D Object-Oriented Visual SLAM with Deblurring \\ for Agile Drones
 \\

}

\author{
  Denis Davletshin,
  Iana Zhura,
  Vladislav Cheremnykh,
  Mikhail Rybiyanov,\\
  Aleksey Fedoseev, and
  Dzmitry Tsetserukou
  % \thanks{$^{1}$Intelligent Space Robotics Laboratory, CDE, Skolkovo Institute of Science and Technology, Moscow, Russia {\tt\small \{Denis.Davletshin, Iana.Zhura, Vladislav Cheremnykh, Mikhail.Rybiyanov, Aleksey.Fedoseev, D.Tsetserukou\}@skoltech.ru}}
  \thanks{The authors are with the Intelligent Space Robotics Laboratory, Center for Digital Engineering, Skolkovo Institute of Science and Technology (Skoltech), 121205 Bolshoy Boulevard 30, bld. 1, Moscow, Russia. { \{denis.davletshin, iana.zhura, vladislav.cheremnykh, mikhail.rybiyanov, aleksey.fedoseev, d.tsetserukou\}@skoltech.ru}}
}

\maketitle

\begin{abstract}
The paper focuses on the algorithm for improving the quality of 3D reconstruction and segmentation in DSP-SLAM by enhancing the RGB image quality. SharpSLAM algorithm developed by us aims to decrease the influence of high dynamic motion on visual object-oriented SLAM through image deblurring, improving all aspects of object-oriented SLAM, including localization, mapping, and object reconstruction. 

The experimental results revealed noticeable improvement in object detection quality, with F-score increased from 82.9\% to 86.2\% due to the higher number of features and corresponding map points. The RMSE of signed distance function has also decreased from 17.2 to 15.4 cm. Furthermore, our solution has enhanced object positioning, with an increase in the IoU from 74.5\% to 75.7\%. SharpSLAM algorithm has the potential to highly improve the quality of 3D reconstruction and segmentation in DSP-SLAM and to impact a wide range of fields, including robotics, autonomous vehicles, and augmented reality.
\end{abstract}

% \begin{IEEEkeywords}
% Visual SLAM, Object Reconstruction, UAV SLAM, Deblurring
% \end{IEEEkeywords}

\section{Introduction}
Among the increased functionality of multi-rotor unmanned aerial vehicles (UAVs), visual perception has become a valuable tool for solving the visual simultaneous localization and mapping (SLAM) problem and obtaining more information about the environment, such as object localization and reconstruction. Given the ability of SLAM to provide accurate and reliable estimates of a vehicle's position and orientation, it has become a popular technique for multi-rotor vehicles \cite{Sharafutdinov2023}. In fact, Visual SLAM allows UAVs to create detailed maps of their surroundings, which can be used for a variety of tasks. 
\begin{figure}[t]
  \centering
 \includegraphics[width=1.0\linewidth]{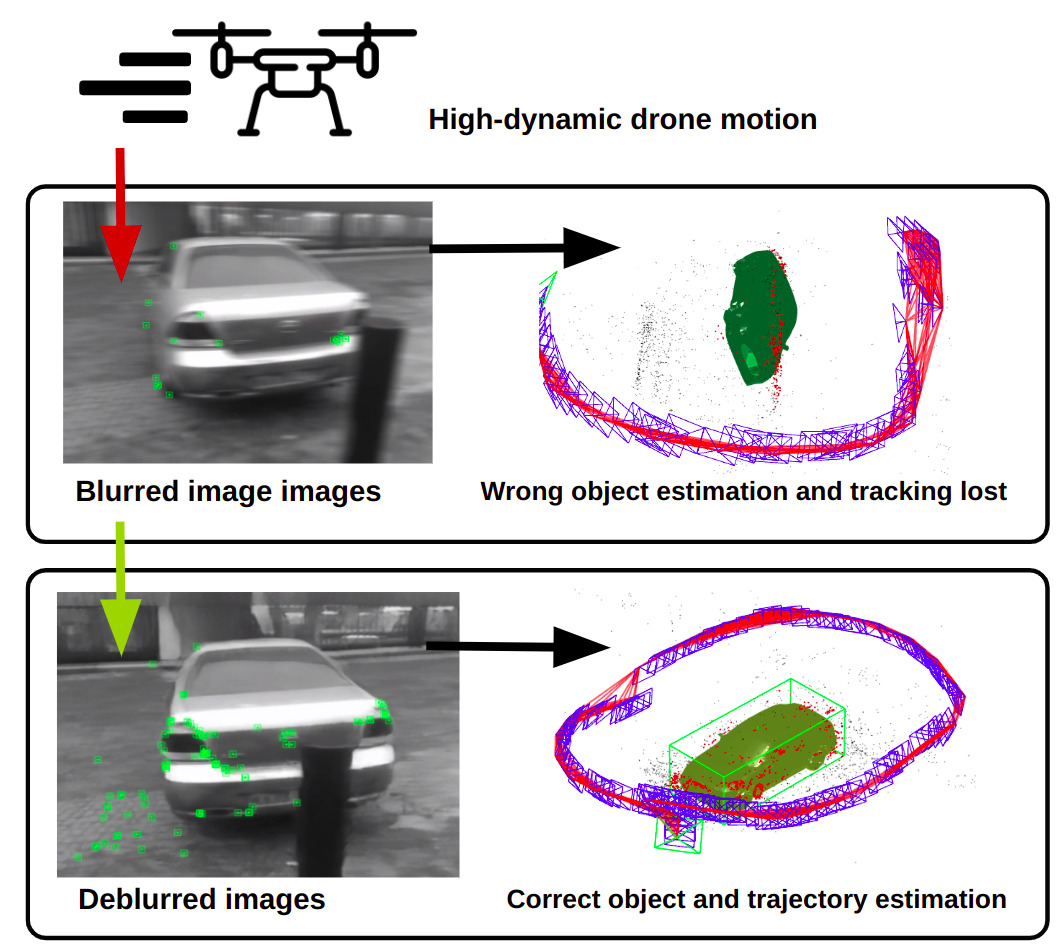}
  \caption{\centering Improvement in visual SLAM performance under the SharpSLAM with image deblurring algorithm.} %Improvment in visual SLAM performance under the SharpSLAM (You can change name ofc but DeblurSLAM occupied) image debluring algorithm
  \label{fig:main_pic}
\end{figure}
Implementation of LiDARs has proven to be efficient in 3D mapping and localization \cite{Niu_2021, Hensel_2022}. However, high cost, as well as sensitivity to reflective surfaces and weather precipitation make them unsuitable for a wide scope of indoor and outdoor applications. The implementation of light-weight and available RGB and RGB-D cameras is therefore highly demanded for a robot localization with visual SLAM \cite{Kazerouni_2022, Kruzhkov2022}. However, even with the advances of the solutions based on NeRF \cite{Nenashev2023}, cameras without depth information until recent years presented a high challenge for accurate mapping. In addition, visual feedback from the camera is often exposed to several drawbacks caused by bad lighting conditions or lack of textures on surfaces. Several algorithms were developed to overcome these challenges, e.g., DarkSLAM \cite{Savinykh_2022} for dim lighting and Pop-up SLAM \cite{Yang_2016} for textureless environments.

One of the biggest challenges for RGB-based visual SLAM remains in the unpredictable blurriness of the images derived from the camera. While several control \cite{Li_2015} and data processing \cite{Yeh_2022} approaches were developed to compensate unwanted camera oscillation, the image blurriness still affects high-speed dynamic systems, such as UAVs. This issue especially affects the application of drones in 3D reconstruction of buildings and other large-scale objects \cite{Filatov_2021, Noda_2022} limiting the scope of applications for UAVs carrying RGB sensors.

To address this issue, we propose a novel SharpSLAM algorithm that utilizes a generative adversarial networks (GAN) architecture to deblur low-quality image frames and improve the precision of the environment reconstruction. The main contribution of this work is the development and evaluation of the algorithm for both localization and reconstruction suitable for highly dynamic UAV systems.

\section{Related Works}

\subsection{Visual SLAM}

Visual SLAM is particularly useful for UAV tasks because of it is light-weight, cost-effective, and precise compared to other mapping techniques, such as Light Detection and Ranging (LiDAR).
Several algorithms were developed for visual SLAM, for example, ORB-SLAM2 \cite{Mur-Artal_2017} and most recent ORB-SLAM3 \cite{Campos_2021} achieving a high-precision result with a wide scope of monocular, stereo, and RGB-D cameras. However, the proposed algorithms still have challenges in the accuracy and completeness of 3D reconstruction, which can be caused by factors such as lighting conditions, camera calibration, and motion blur. This can lead to errors in the UAV's perception of the environment, which can impact its ability to navigate and perform tasks effectively.

\subsection{Object-oriented SLAM}

The simultaneous solution of both the SLAM and the deblurring problem is a useful approach for dynamic systems. One of the proposed methods to achieve it is to restore images using fast deconvolution and SLAM data. Additional features are then extracted and registered to the map, allowing SLAM to continue even with the blurred images \cite{Lee-2011}. The aforementioned SLAM system has been validated on the TUM RGB-D dataset and used in mobile manipulators for applications involving object grasping with high accuracy for low-speed robotic arms \cite{Peng-2019}.

% Another approach suggested in DeepSDF algorithm \cite{Park_2019} is learning the continuous signed distance functions to achieve precise shape representation of the reconstructed objects. The suggested approach achieved an efficient feature extraction in different lighting conditions, however, it requires extensive computation power for training and its scalability with objects and areas might require further evaluation.

The inspiration of this work is the Object Oriented SLAM with Deep Shape Priors (DSP-SLAM) \cite{DSP-SLAM}, which demonstrated full object reconstructions of high quality even from incomplete observations while maintaining a reliable global map. Evaluations of this algorithm also revealed decreased camera tracking drift and improvements in object pose and shape reconstruction compared to recent deep prior-based reconstruction techniques.

Recently, with the rise of deep learning techniques, significant improvements have been made in image quality enhancement, as discussed in \cite{dynamicblur} and \cite{Savinykh_2022}. Among these techniques, single motion deblurring based on deep learning approaches has been introduced in \cite{chen2022motion}. Although, it is worth noting that a more flexible and efficient approach to enhancing image quality is to utilize end-to-end GANs and the Feature Pyramid Network, which enable faster processing results, as discussed in \cite{DeblurGAN-V2}.

It is clear that deblurring plays a crucial role in improving the accuracy of SLAM and 3D segmentation, and existing techniques have their limitations. Utilizing GANs for end-to-end image quality enhancement offers a promising avenue for future research in this area. Nevertheless, it is important to note that for UAV systems, it is not only important to accurately perform SLAM but also to accurately reconstruct 3D environments and segment objects within them. Therefore, it is crucial to analyze 3D segmentation after deblurring, as this can significantly impact the ability of UAV systems to define objects in complex environments.

\subsection{3D Reconstruction for UAV}
Obtaining 3D models of physical buildings and landscapes is often desirable for a better understanding of the robot's surroundings. Recently, research has been conducted to explore the possibility of using 3D reconstruction and segmentation for UAV tasks. However, many of these techniques rely on LiDAR data, which can be sparse and expensive \cite{to2021drone}. In contrast, Visual SLAM is widely used for UAV perception tasks in complex environments and the attempts to tackle the limitations of Visual SLAM for UAV purposes have been done \cite{rs12203308}, \cite{drones6030079}. Additionally, a novel concept of 3D reconstruction by heterogeneous swarm was proposed and is relying on visual data from on-board cameras \cite{Zhura2023}. Nevertheless, the computational complexity and time consumption have been addressed.

One of the primary challenges in visual SLAM is mitigating limitations caused by harsh environmental conditions and motion blur, which have been previously considered. Several works explore this issue, e.g., the quality assessment approach for 3D reconstruction, suggested in \cite{Dronova2024}. However, current techniques have not fully addressed the issue of 3D segmentation of areas of interest while simultaneously addressing the limitations of visual SLAM.

\subsection{Deblurring techniques in SLAM and object detection}
To enhance the accuracy of SLAM and segmentation, a common approach is to preprocess the input images. In UAV systems, images are often blurred, which can result in a lack of interest points and inaccuracy of their. To address this issue, deblurring algorithms have been integrated into SLAM, showing superior performance compared to standard Visual SLAM baselines, as demonstrated in \cite{Guo_2021}. However, it is important to note that the impact of deblurring algorithms on 3D reconstruction, when combined with SLAM, remains largely unexplored.

\section{System Overview: Hardware and Software}

\begin{figure}[ht!]
  \centering
%       \framebox{\parbox{3in}{
% }}
 % \includegraphics[scale=0.23]{images/multiagent.jpg}
 \includegraphics[width=1\linewidth]{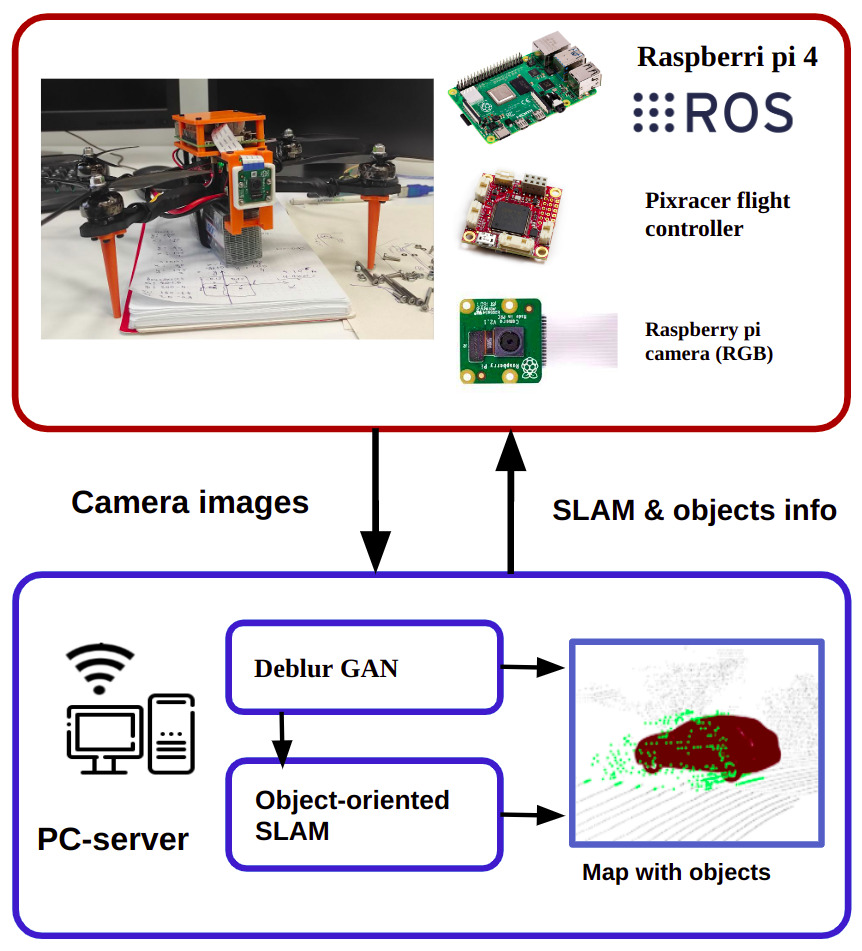}
  \caption{The overview of SharpSLAM hardware architecture used for the experimental evaluation of the proposed algorithm. Arrows show direction of data exchange between modules.}
  \label{fig:hardware}
\end{figure}

SharpSLAM hardware has two components: a UAV that takes pictures and sends them to the stationary server, which is responsible for deblurring images and processing SLAM with object reconstruction (see Fig. \ref{fig:hardware}). Once the server has finished, it sends the map and object data back to the drone. The drone is equipped with Raspberry Pi 4 with 8GB of RAM and a Raspberry Pi camera v2.1 (8 MP Sony IMX219 image sensor). We selected Pixracer R14 to serve as the flight controller for the UAV, because it provides all functionality to communicate via ROS protocols.

\begin{figure}[ht!]
  \centering
%       \framebox{\parbox{3in}{
% }}
 % \includegraphics[scale=0.23]{images/multiagent.jpg}
 \includegraphics[width=1\linewidth]{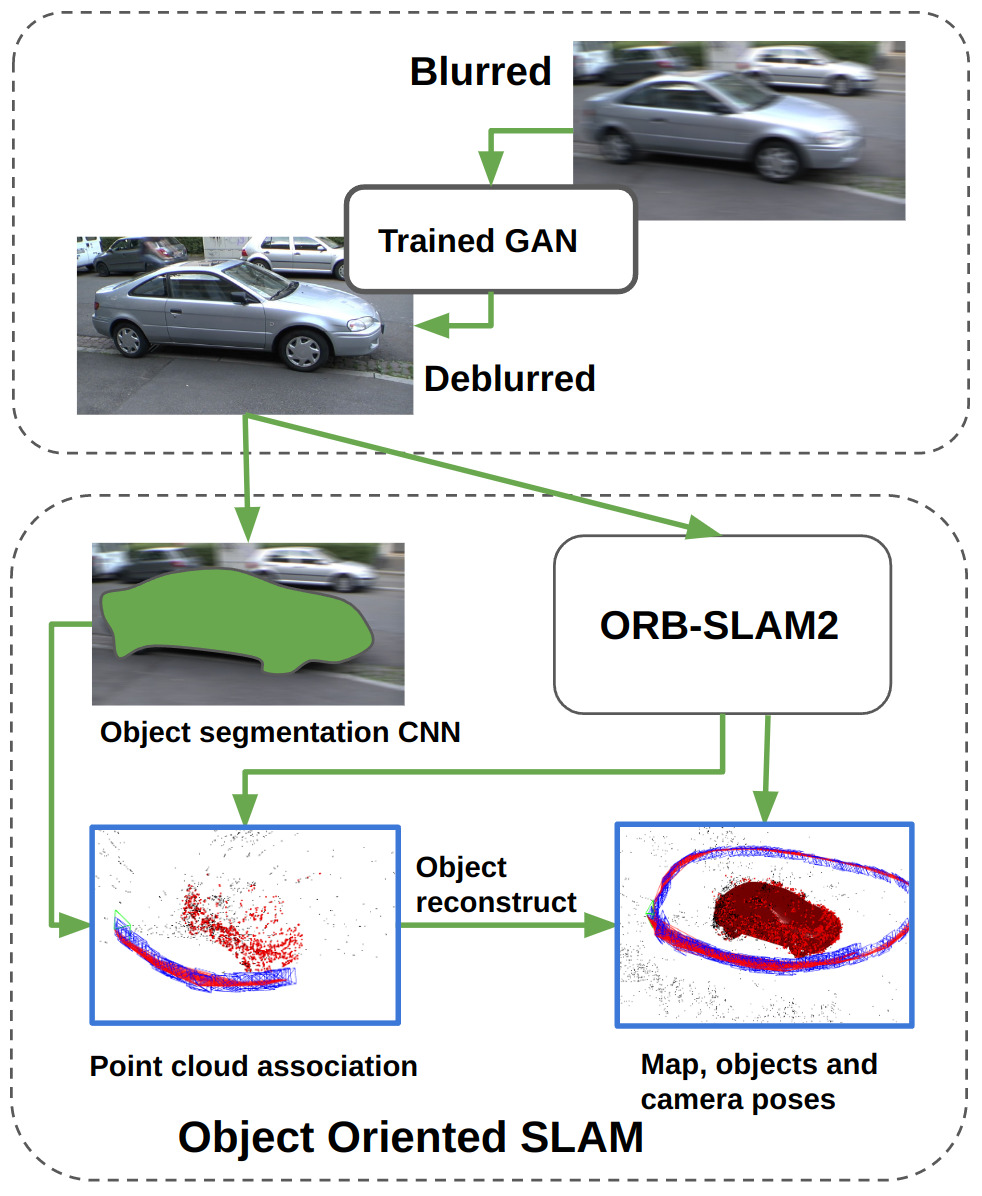}
  \caption{Visual SLAM pipeline with SharpSLAM approach.}
  \label{fig:software}
\end{figure}

The Software consist of three blocks (see Fig. \ref{fig:software}). The first block is the deblurring algorithm, which receives images from the drone and reduces the degree of motion blur. The second one is the visual SLAM based on ORB-SLAM2 \cite{Mur-Artal_2017}, which takes pictures after deblurring, provides a point cloud and takes part in robot localization. In parallel, images are also sent to the third block, which is responsible for object reconstruction. It uses 2D object segmentation from images and point cloud information to find the 3D shape and 6-DoF pose of the objects.

\subsection{Object-oriented SLAM}
DSP-SLAM \cite{DSP-SLAM} was chosen to serve as the object-oriented SLAM algorithm. This system applies 2D object segmentation on keyframes taken from ORB-SLAM 2, and provides map point association with objects. The process of shape reconstruction is iterative optimization. Each object is represented as a vector {$\mathbf{z}$} that will be iteratively optimized. Map points are then processed with the DeepSDF \cite{Park_2019_CVPR} neural network decoder that returns SDF for specific type of object, and it represents not fixed shape, but an entire class of shapes, such as hatchback, coupe, sedan etc. (if they were present in training datasets). DSP-SLAM algorithm utilize this decoder to optimize the following sequence:

\begin{equation}
    E = \lambda_s E_{surf} + \lambda_r E_{rend} + \lambda_c ||z||^2
\end{equation}

The first term takes into account a consistency between associated point cloud and reconstructed surface. The second term is a rendering loss, that enables to reconstruct object correctly even if a case of partially observed object. Also it this term is sensetive to growth of shape outside segmentation area. The third is a shape code regularization term. The output object poses are added to the optimization graph of the SLAM, so the optimization of object reconstruction affects the solution of the SLAM problem and vice versa. 

\subsection{Deblurring with Neural Network}

To remove motion blur from the UAV, we started with the DeblurGANv2 neural network. The Feature Pyramid Network is employed by this network. In our experiment, we made use of the Inception-ResNet-v2 architecture. MobileNet architecture should be used for real-time deblur applications as it has a faster inference time than ResNet architecture \cite{DeblurGAN-V2}.

\section{Experiments}
\subsection{Dataset preparation}\label{A}

% In order to reproduce conditions of high dynamic motion we applied an algorithm which imitates blurring in a realistic way. \\

% \subsubsection{Dataset collection}\label{AA} 
% The dataset was collected considering two different scenarios. The first scenario was to move around an object of interest, keeping the object in the center of the camera frame. The second scenario was to move through a parking lot filled with cars, where the cars were in the frame for a shorter period of time than in the first scenario. 

The dataset for this project was collected in the parking garage of Skoltech University, featuring a Kia Soul car as the subject. The drone was flown in an elongated ring loop around the car by the drone operator, ensuring that the car was kept within the center of the camera frame. Each recording lasted for 3 minutes, recorded in a resolution of 1980x1280 with a frame rate of 15 frames per second. 

\subsection{DSP-SLAM parameters}

\begin{figure}[ht!]
 \centering
% \framebox{\parbox{3in}{
% }}
 \includegraphics[width=1\linewidth]{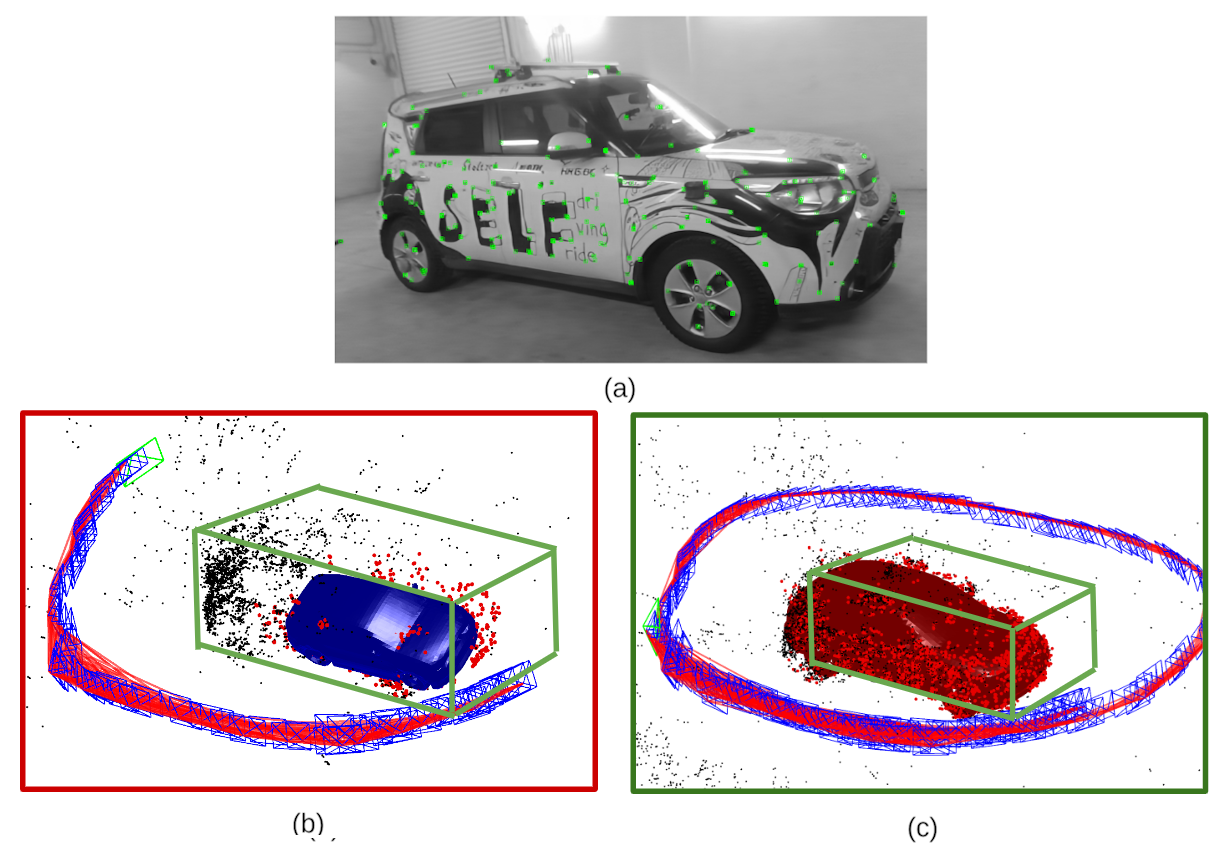}
 \caption{Results of DSP-SLAM reconstruction for different number of keyframes waited before start reconstruction. (a) Initial camera position. (b) No. keyframes = 15, reconstruction was too early. (c) No. keyframes = 50, improved reconstruction after observing all sides of the car}
 \label{fig:dsp-adjust}
\end{figure}

At the first iterations of SharpSLAM verification, we discovered that in case of cars with the shape that significantly differs from sedan configuration, DSP-SLAM may reconstruct its shape not correctly. Thus, in different scenarios of taking dataset we need to adjust DSP-SLAM algorithm. One of the most critical parameters was a number of keyframes for reconstruction initialization. The typical example of inaccurate reconstruction scenario is a start of car observation from the side. In that case if DSP-SLAM starts to find the shape too early (too small number of keyframes), the algorithm will have a point cloud only from one side, and these points probably not be enough. In the Fig. \ref{fig:dsp-adjust}(b) was shown that early attempt of reconstruction causes wrong position and dimension estimation. At the Fig. \ref{fig:dsp-adjust}(c), on the contrary, was shown that increased number of keyframes can improve the object pose estimation. \\

Empirically there was found an optimal parameter of 50 keyframes that is sufficient to observe the major part of the vehicle. 

\subsection{Monocular SLAM position and scale calibration}
\begin{figure}[ht!]
  \centering
%       \framebox{\parbox{3in}{
% }}
 % \includegraphics[scale=0.23]{images/multiagent.jpg}
 \includegraphics[width=1\linewidth]{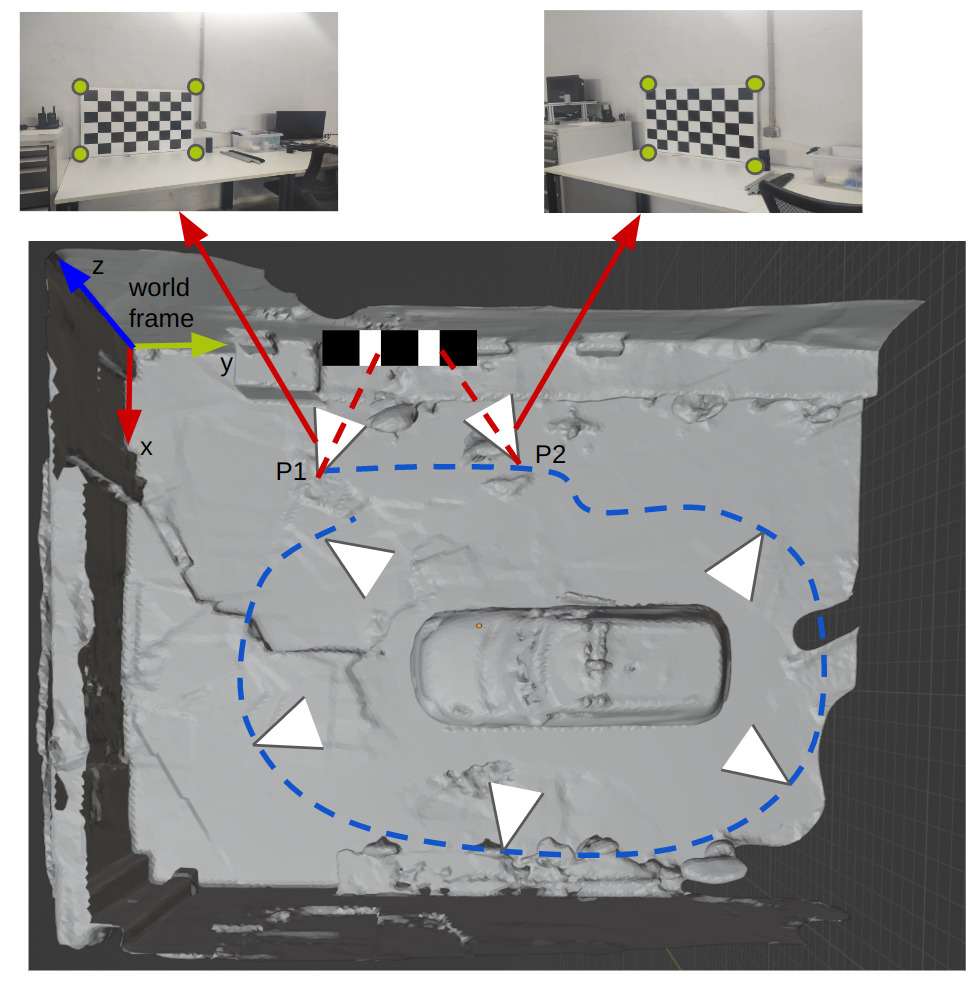}
  \caption{Schematic explanation of SLAM scale and position calibration.}
  \label{fig:calib_scheme}
\end{figure}

In order to obtain quantitative results for object reconstruction and positioning, we need to have a map of our SLAM and ground truth in the same coordinate system. Therefore, we performed calibration of the scale and coordinate system with the help of a calibration chessboard Fig. \ref{fig:calib_scheme}. To get scale information, we find positions of 2 frames (P1 and P2) relative to the board, and relative to the SLAM coordinate system. After that we can simply get scale according to the following equation:

\begin{equation} \label{eq:scale}
    s = \frac{\rho(P_{1cb},P_{2cb})}{\rho(P_{1cs},P_{2cs})}
\end{equation}

\noindent
 where $\rho$ is the Euclidean distance between 3D points, $P_{cb}$ is the position of camera with respect to chessboard, and $P_{cs}$ is the position of camera in SLAM coordinate system.

After scale calibration we perform coordinate system calibration. We detected camera position relative to the board at the frame $P_1$ by solving Perspective-n-Point problem with OpenCV library. The position of chessboard relative to the world is known, therefore we can obtain transformation between SLAM and world coordinate systems.

% \subsection{Camera localization validation}\label{AA}
% To validate the correctness of the trajectory, we used a stereo visual SLAM integrated in an Intel RealSense t265, which was considered a ground truth trajectory. Then we compared it with the trajectories from visual odometry using blurred and restored images captured by the monocular Raspberry Pi camera V2.1. 

% \begin{table}[!ht]
% \caption{\centering Overall odometry errors.}
% \label{table_metrics}
% \begin{center}
% \begin{tabular}{|c|c|c|c|}
% \hline
% Dataset & Distance traveled, m & Translation error, \% & Rotation error deg/m \\
% \hline
% % Raw & \textbf 0.997 & 0.834 & \textbf 0.908\\
% \hline
% Blurred & \textbf 0 & 0 & \textbf 0 \\
% \hline
% Deblurred & \textbf 0 & 0 & \textbf 0\\

% % \hline
% % \hline
% % Dataset & № of points &  RMSE SDF, m & IoU, \% \\
% % \hline
% % % Raw & \textbf{30} & 0.124 & \textbf 83.19 \\
% % \hline
% % Blurred &  0 & 0 & 0 \\
% % \hline
% % Deblurred & 0 & 0 &  0 \\

% \hline
% \end{tabular}
% \end{center}
% \end{table}

% \subsection{Recognition success}\label{AA}

\subsection{Shape reconstruction validation}\label{AA}
\begin{figure}[ht!]
  \centering
%       \framebox{\parbox{3in}{
% }}
 % \includegraphics[scale=0.23]{images/multiagent.jpg}
 \includegraphics[width=1\linewidth]{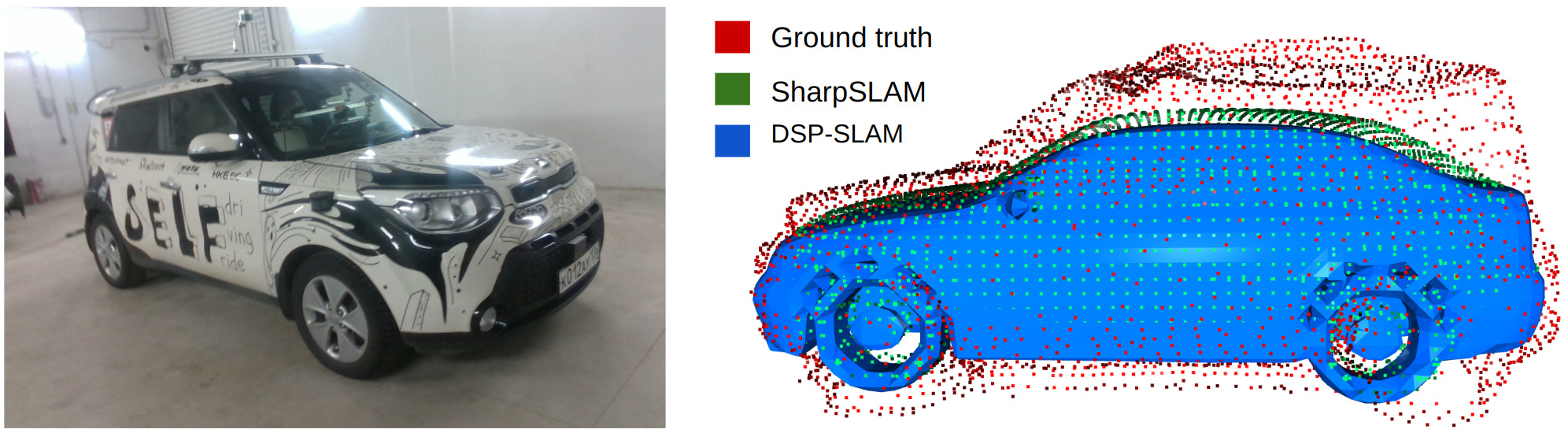}
  \caption{\centering Comparison of qualitative results relative to the ground truth.}
  \label{fig:shape_compare}
\end{figure}

To estimate the quality of the 3D segmented model obtained from the SharpSLAM, 
reconstructed 3D models was compared with LiDAR point cloud as a ground truth car model. The intersection over union (IoU) of 3D bounding boxes and root mean square error (RMSE) of the signed distance function (SDF) over all predicted mesh points have been calculated. According to the Table \ref{table_metrics}, our approach improves the quality of object reconstruction. In Fig. \ref{fig:shape_compare} it can be seen that the roof of mesh car made with Sharp-SLAM is closer to the ground truth than the same one from DSP-SLAM. 

\begin{table}[!ht]
\caption{Geometry Metrics of Reconstructed Objects.}
\label{table_metrics}
\begin{center}
\begin{tabular}{|c|c|c|c|}
\hline
Dataset & Precision & Recall & F-score \\
\hline
% Raw & \textbf 0.997 & 0.834 & \textbf 0.908\\
\hline
Blurred & 0.971 & 0.723 & \textbf {0.829}\\
\hline
Deblurred & 0.987 & 0.764 & \textbf {0.862}\\

\hline
\hline
Dataset & No. of points &  RMSE SDF, m & IoU, \% \\
\hline
% Raw & \textbf{30} & 0.124 & \textbf 83.19 \\
\hline
Blurred & 2337 & 0.172 & \textbf {74.51} \\
\hline
Deblurred & 3236 & 0.154 & \textbf {75.67} \\

\hline
\end{tabular}
\end{center}
\end{table}

% OLD VERSION
% The experiment revealed that the number of points associated with the car increased by 38.46\% after the deblurring processing. This was mostly caused by the tracking loss in case of blurred images. Increased number of map points also improves the positioning guess and leads to more reliable 3D reconstruction behavior. The recall value of segmentation (car detection) increased by 4.1\% and the F-score by 3.3\% as a result of these improvements. Also worth noting is that while SLAM behaved unpredictably when fed datasets with blurry images and frequently lost the camera trajectory, interfering with the process of creating a map of the area. After image processing with neural networks, these issues were not seen on datasets with restored images.

The experiment showed a 38.46\% increase in the number of points associated with the car after the deblurring process. This improvement was mainly due to reduced tracking loss in cases of blurred images. The higher number of map points also enhanced the accuracy of the initial position estimate, leading to more reliable 3D reconstruction. Additionally, the recall for car segmentation improved by 4.1\%, and the F-score increased by 3.3\% as a result of these enhancements. It is important to note that SLAM performed unpredictably when processing datasets with blurry images, often losing track of the camera’s trajectory and interrupting the mapping process. However, after image restoration using neural networks, these issues were no longer present in the datasets with enhanced images.

\section{Conclusion and Future Work}

In this work a new visual SLAM algorithm was proposed for UAV localization and object reconstruction. The proposed approach is based on DSP-SLAM with image enhancement achieved using the DeblurGAN-V2 neural network with the MobileNet architecture, allowing UAVs to perform real-time image restoration based on RGB cameras only.

% 3D segmented models was compared for different cases: 

An experiment was carried out to evaluate the improvement of the SLAM algorithm. The results revealed that, with the number of detected points corresponding to the car increased by 38,46\%, the IoU showed an increase from 74.51\% to 75.67\%. The RMSE of the signed distance function for the trajectory decreased from 17.2 cm to 15.4 cm. The obtained results suggested that the proposed SharpSLAM approach may potentially increase the precision of object reconstruction for UAVs and other autonomous systems with a high-speed motion. In the future we are planning to optimize the SharpSLAM reconstruction pipeline to improve the initial position estimation and correction of object estimation while receiving new data. Also we will conduct additional experiments with a multi-agent SLAM system running on several UAVs as they potentially enhance coverage of observing area and hence reconstruction quality. Furthermore, different reconstruction strategies will be developed and evaluated with several objects of different size. 

\section*{Acknowledgements} 
Research reported in this publication was financially supported by the RSF grant No. 24-41-02039.

\bibliographystyle{IEEEtran}
\bibliography{References.bib}

\vspace{12pt}

\end{document}